\documentclass[letterpaper, 10 pt, conference]{ieeeconf}  

\IEEEoverridecommandlockouts                              

\usepackage{hyperref}
\usepackage{amsmath}
\usepackage{amssymb}
\usepackage{multirow}
\usepackage{booktabs}
\usepackage{tikz}
\usepackage{array}
\usepackage{pifont}  
\usepackage{colortbl}  
\usepackage{xcolor}  
\usepackage{balance}
\usepackage{float}
\usepackage{placeins}
\usepackage{xcolor}

\title{\LARGE \bf
RTAGrasp: Learning Task-Oriented Grasping from Human Videos via Retrieval, Transfer, and Alignment
}

\author{Wenlong Dong$^{1,2}$, Dehao Huang$^{1,2}$, Jiangshan Liu$^{2}$, Chao Tang$^{1,2}$ and Hong Zhang$^{1,2}$ \emph{Fellow, IEEE}
\thanks{$^{1}$Shenzhen Key Laboratory of Robotics and Computer Vision, Southern
University of Science and Technology, Shenzhen, China.}%
\thanks{$^{2}$Department of Electronic and Electrical Engineering, Southern University of Science and Technology, Shenzhen, China. {\tt \small dongwl2023@mail.sustech.edu.cn}}%
}

\begin{document}

\maketitle
\thispagestyle{empty}
\pagestyle{empty}

\begin{abstract}

Task-oriented grasping (TOG) is crucial for robots to accomplish manipulation tasks, requiring the determination of TOG positions and directions. Existing methods either rely on costly manual TOG annotations or only extract coarse grasping positions or regions from human demonstrations, limiting their practicality in real-world applications. To address these limitations, we introduce RTAGrasp, a Retrieval, Transfer, and Alignment framework inspired by human grasping strategies. Specifically, our approach first effortlessly constructs a robot memory from human grasping demonstration videos, extracting both TOG position and direction constraints. Then, given a task instruction and a visual observation of the target object, RTAGrasp retrieves the most similar human grasping experience from its memory and leverages semantic matching capabilities of vision foundation models to transfer the TOG constraints to the target object in a training-free manner. Finally, RTAGrasp aligns the transferred TOG constraints with the robot's action for execution. Evaluations on the public TOG benchmark, TaskGrasp dataset, show the competitive performance of RTAGrasp on both seen and unseen object categories compared to existing baseline methods. Real-world experiments further validate its effectiveness on a robotic arm. Our code, appendix, and video are available at \url{https://sites.google.com/view/rtagrasp/home}.
\end{abstract}

\section{Introduction}

\begin{figure}[t]
  \centering
  \vspace*{0.08in}
  \begin{tikzpicture}[inner sep = 0pt, outer sep = 0pt]
    \node[anchor=south west] (fnC) at (0in,0in)
      {\includegraphics[width=\linewidth,clip=true]{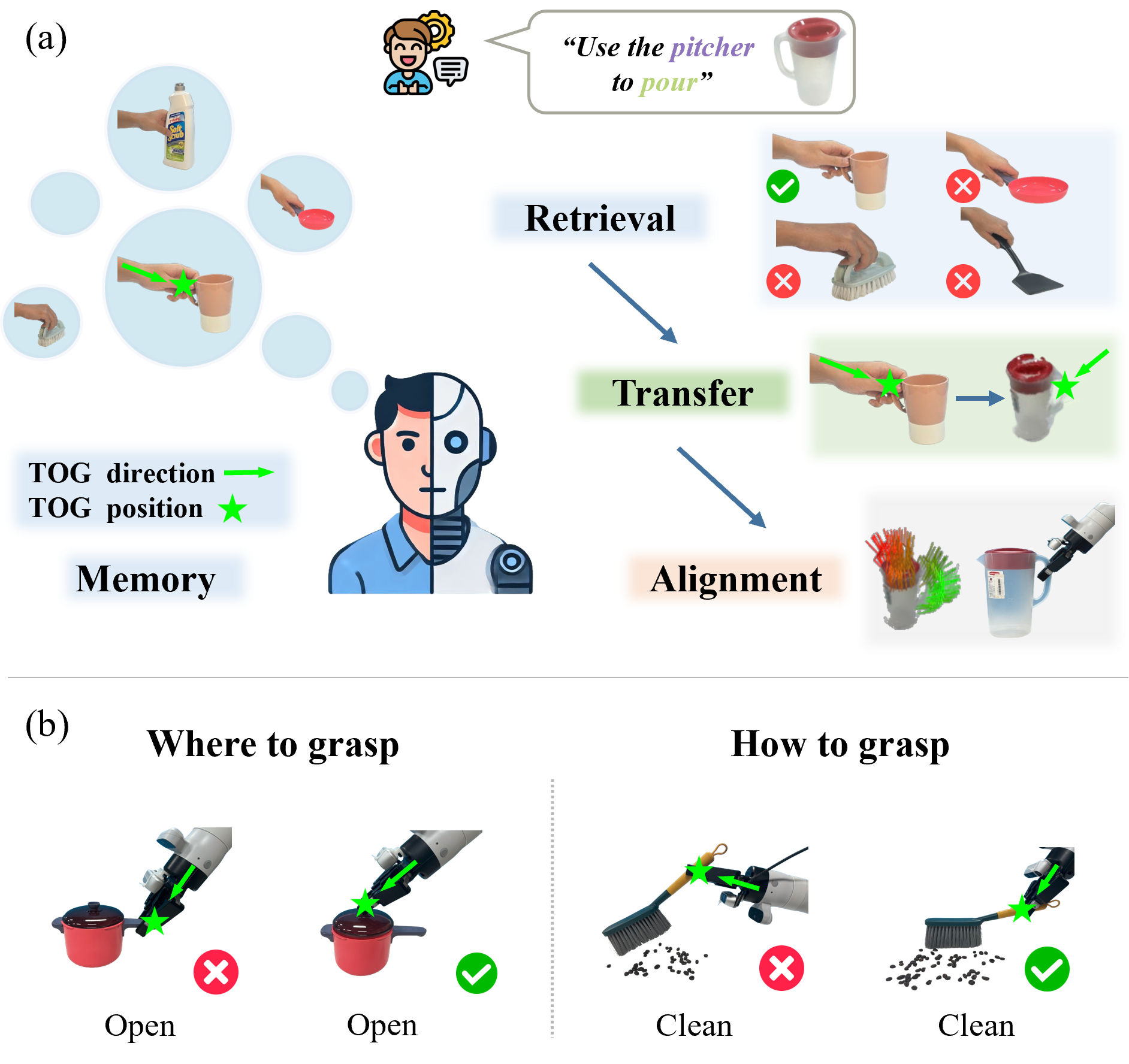}};
  \end{tikzpicture}
  \vspace*{-0.2in}
  \caption{(a) Robots learn TOG skills from human demonstration videos through Retrieval, Transfer, and Alignment. (b) An incompatible TOG position or direction could result in the failure to complete subsequent tasks.} 
  \label{fig:intro}
  \vspace*{-0.2in}
\end{figure}

To successfully manipulate an object and complete subsequent household tasks, a robot must first grasp the object in a task-oriented manner. As shown in Fig. \ref{fig:intro}(b), the robot needs to determine an optimal grasping position related to the task (i.e., ``where to grasp") and a grasping direction that is compatible with the task (i.e., ``how to grasp"). Failure to consider either grasping constraint would result in unsuccessful completion of subsequent manipulation tasks.

To learn such TOG skills, training-based methods typically require the collection and manual annotation of TOG datasets covering a variety of tasks and objects, which are then used for training TOG models. For example, GraspGPT \cite{tang2023graspgpt} relies on a manually annotated dataset to establish semantic relationships between objects in the dataset and novel ones. While training-based methods \cite{tang2023graspgpt, wei2024grasp, murali2021taskgrasp} have achieved some success, TOG data collection and annotation are costly and labor-intensive. Additionally, the scarcity of annotated TOG data challenges the ability of training-based methods to generalize to unseen objects and tasks.

On the other hand, recent works directly utilize the common sense knowledge from foundation models to predict task-oriented grasps in a training-free manner. LERF-TOGO \cite{rashid2023LERFTOGO} constructs a language-embedded radiance field using features extracted from VLMs to perform zero-shot semantic grasping. Lan-grasp \cite{mirjalili2023langrasp} prompts an LLM to output the names of object parts suitable for grasping, and then uses a VLM to identify these parts in images. RoboABC \cite{ju2024roboABC} uses CLIP \cite{radford2021CLIP} and Stable Diffusion \cite{rombach2022SIFT} for contact point retrieval and transfer. With the open-ended knowledge from foundation models, training-free methods avoid costly TOG data collection and annotation. However, foundation models can only provide coarse prior knowledge of grasping positions or regions, which is insufficient for determining task-oriented grasps.

Cognitive psychology research \cite{bobrowicz2020flexibility} indicates that when humans learn a new tool manipulation skill, they often apply previous experiences, regarding where and how to manipulate a tool, through analogical transfer. Notably, we have found that human demonstration videos can provide such experiences for TOG, containing complete TOG constraints (i.e., positions and directions), with minimal collection and annotation effort. Based on this insight, we propose RTAGrasp, a Retrieval, Transfer, and Alignment framework to effortlessly extract complete TOG constraints from human demonstration videos and transfer them to robots for execution.

As illustrated in Fig. \ref{fig:intro}(a), our approach begins by automatically processing human grasping demonstration videos to extract complete TOG constraints for various objects and tasks, which are then used to construct a robot memory. Given a task instruction and a visual input of the target object, RTAGrasp retrieves the most relevant TOG experience from its memory, analyzing both semantic and geometric similarities. The retrieved TOG constraints are then transferred to the target object using vision foundation models. Finally, RTAGrasp aligns the transferred TOG constraints to the robot's action for execution. Evaluations on the public TOG benchmark, TaskGrasp dataset, demonstrate that RTAGrasp outperforms existing baselines in both seen and unseen object categories. Furthermore, we deploy RTAGrasp on a Kinova Gen3 robotic arm to validate its effectiveness in real-world applications.

\section{Related Works}

The core objective of TOG is to jointly address both ``where to grasp" and ``how to grasp". Classical TOG methods \cite{haschke2005taskold1} \cite{prats2007taskold2} evaluate task-oriented grasp quality through task wrench space analysis. However, a significant limitation of this approach is its reliance on complete object and hand models, which are challenging to collect or reconstruct. In recent years, deep learning-based methods have made progress in addressing the TOG problem, which can be categorized into training-based and training-free methods.

\subsection{Training-based Methods}

Training-based methods typically require the creation of manually annotated TOG datasets for training. For example, Murali et al. \cite{murali2021taskgrasp} collect the largest and most diverse TOG dataset, which is then used to train a task-oriented grasp evaluation network. GraspCLIP \cite{tang2023task} trains an end-to-end, vision-language task-oriented grasp prediction model using a manually annotated dataset. Jin et al. \cite{jin2024reasoning} propose generating task-oriented grasp poses using a multimodal large language model, which depends on manually annotated datasets for training. Nguyen et al. \cite{nguyen2024language} contribute a 3D Affordance-Pose (3DAP) dataset to jointly address open-vocabulary affordance detection and task-oriented grasp pose generation. GraspGPT \cite{tang2023graspgpt} leverages the prior knowledge of large language models to establish semantic relationships between objects in the dataset and unseen objects, achieving the state-of-the-art (SOTA) results. Although training-based methods have achieved some degree of success, they heavily depend on labor-intensive manual TOG annotations. Additionally, the scarcity of annotated TOG data limits their generalization to unseen objects. In contrast, our approach extracts TOG constraints from human demonstration videos without any manual effort and explicitly transfers them to robot's actions using foundation models.

\subsection{Training-free Methods}

Most training-free methods leverage foundational models \cite{mirjalili2023langrasp}, \cite{li2024shapegrasp}, \cite{huang2024efficient}, \cite{ge2024}, pre-trained on large-scale internet data, to effectively map embedded semantic knowledge to target objects and select the optimal TOG position or region based on task specifications. For instance, based on the commonsense reasoning capabilities of LLM, ShapeGrasp \cite{li2024shapegrasp} achieves zero-shot TOG for novel objects through geometric decomposition. LERF-TOGO \cite{rashid2023LERFTOGO} combines VLMs with 3D scene reconstruction, enabling robots to accurately grasp specific parts of objects based on natural language queries. Lan-grasp \cite{mirjalili2023langrasp} determines the TOG part of the object using an LLM, and then uses a VLM to ground that part in the image. These methods fully exploit the common sense reasoning capabilities of foundation models, avoiding model training and TOG data collection/annotation. However, foundation models are only able to provide coarse prior knowledge of TOG position or region constraints, which is insufficient to determine a precise task-oriented grasp. In contrast, our approach extracts both TOG position and direction constraints from human demonstration videos and uses semantic correspondences to transfer them to the target object. This effectively addresses both ``where to grasp" and ``how to grasp" challenges.

The work most similar to ours is RoboABC \cite{ju2024roboABC}, which uses CLIP \cite{radford2021CLIP} and Stable Diffusion \cite{rombach2022SIFT} for contact point retrieval and transfer. However, it lacks the ability to flexibly select appropriate grasping positions for different tasks (i.e., TOG positions) and does not address the problem of ``how to grasp" (i.e., TOG directions).

\section{Method}

\begin{figure*}[th]
  \centering
  \begin{tikzpicture}[inner sep = 0pt, outer sep = 0pt]
    \node[anchor=south west] (fnC) at (0in,0in)
      {\includegraphics[height=3.5in,clip=true,trim=0in 0in 0in 0in]{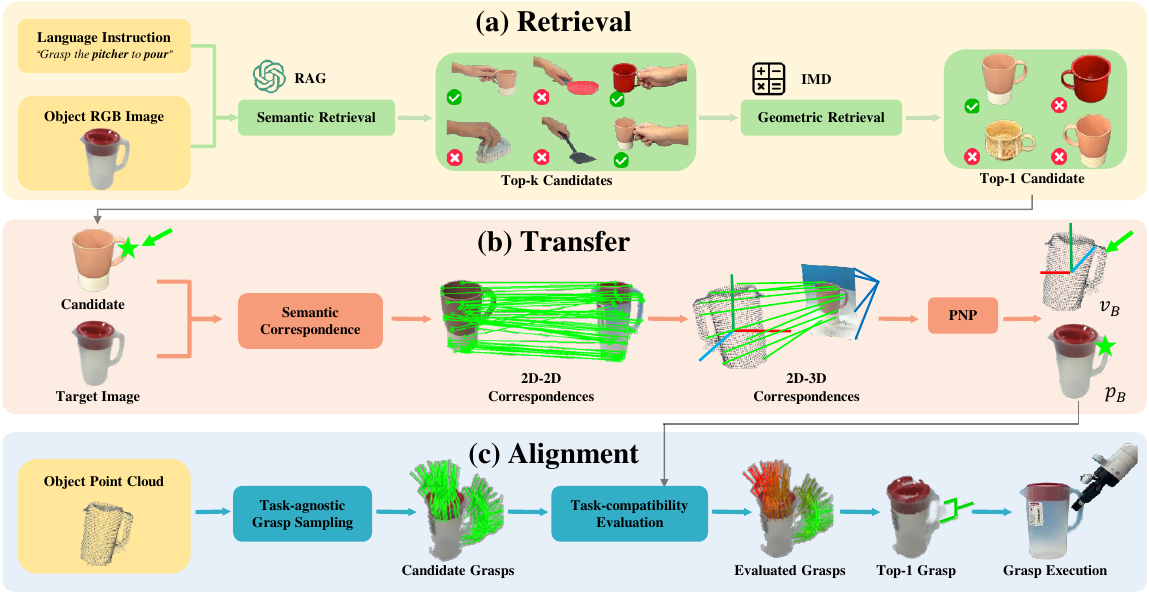}};
  \end{tikzpicture}

    \caption{Overview: the pipeline first utilizes (a) a retrieval module to retrieve the optimal grasping experience (i.e., TOG constraints) from the memory. Next, it uses (b) a transfer module to transfer the retrieved TOG constraints to the target object to obtain the TOG position  $\mathbf{p}_{B}$ and the TOG direction $\mathbf{v}_{B}$. Finally, (c) an alignment module aligns the transferred TOG constraints to the robot's action for execution.}
  \label{fig:pipeline}
  \vspace*{-0.1in}
\end{figure*}

\subsection{Overview}

The core objective is to (1) extract complete TOG constraints from human demonstration videos without manual effort and (2) effectively transfer and align human grasping experiences to target objects for robot grasping. To achieve this, we propose a four-stage pipeline, including memory construction, retrieval, transfer, and alignment. The proposed approach begins by extracting complete TOG constraints from RGB human demonstration videos for a set of objects and tasks, constructing a TOG robot memory. Next, as shown in Fig. \ref{fig:pipeline}, when given an RGB-D image of the target object and a task instruction, RTAGrasp employs semantic and geometric retrieval strategies to retrieve the most similar grasping experience from the memory. Then, RTAGrasp transfers the retrieved TOG constraints to the target object with vision foundation models by establishing 2D-3D correspondence between the retrieved grasping experience and the target object. Finally, to align the transferred TOG constraints to the robot's actions, RTAGrasp samples candidate task-agnostic grasps and evaluates their task compatibility according to the constraints, then selects the optimal grasp for execution.

\subsection{Memory}

\begin{figure}[h]
  \centering
  \vspace*{-0.1in}
  \begin{tikzpicture}[inner sep = 0pt, outer sep = 0pt]
    \node[anchor=south west] (fnC) at (0in,0in)
      {\includegraphics[height=2.9in,clip=true,trim=0in 0in 0in 0in]{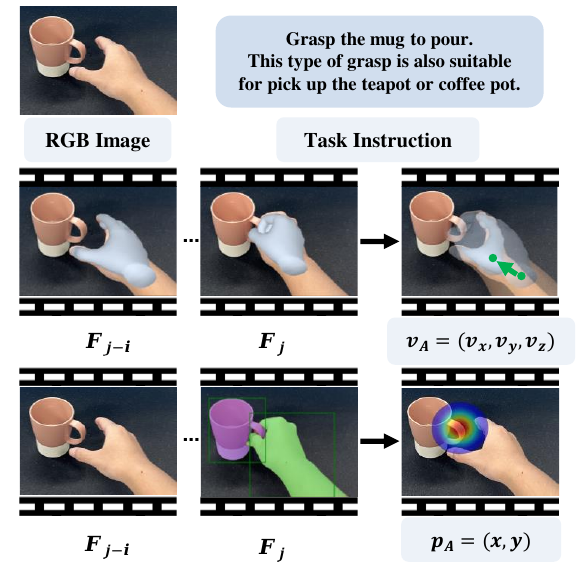}};
  \end{tikzpicture}
    \vspace*{-0.1in}
  \caption{Each instance in the robot memory consists of an object-centered RGB image, a TOG position $\mathbf{p}_A$, a TOG direction $\mathbf{v}_A$, and a task instruction.}
  \label{fig:memory}
  \vspace*{-0.1in}
\end{figure}

We define TOG constraints in the robot memory as TOG positions and directions. Our objective is to extract the information of ``where to grasp" and ``how to grasp" from RGB human demonstration videos. As shown in Fig. \ref{fig:memory}, an instance in the memory consists of four key components: an object-centered RGB image $I_A$, a 2D human TOG position $\mathbf{p}_{A}$ in $I_A$, a 3D human TOG direction $\mathbf{v}_{A}$ in the camera coordinate frame $\mathcal{C}_A$, and a natural language task instruction $T$. Below, we detail our approach to memory construction.

First, we employ a hand-object interaction detector \cite{shan2020handobject} to predict the contact status and extract the bounding boxes of the hand ($B_h$) and the object in contact ($B_o$) from the demonstration video. Once the initial contact frame $F_j$ is identified based on the predicted contact status, we use $B_h$ and $B_o$ to prompt SAM \cite{kirillov2023sam} and extract the contact points between the hand and the object. These contact points are then approximated by a Gaussian distribution, with the TOG position $\mathbf{p}_A$ derived as the Gaussian mean.

Next, we use an off-the-shelf hand reconstruction model \cite{pavlakos2024hamer} to obtain the 3D TOG direction in $\mathcal{C}_A$. Previous research \cite{cheng2021graspsvariedobjectorientation} indicates that the grasping direction of the human hand is consistent with the approaching direction to the object. Therefore, we reconstruct the hand in the 20 frames preceding $F_j$, obtaining the 3D coordinates in $\mathcal{C}_A$ of the wrist joint in each frame and calculating the direction in which the hand approaches the object. During this period, we assume that the camera remains stationary and thus we adopt this direction as $\mathbf{v}_{A}$. To provide diverse viewpoints in the memory, we further augment $I_A$ with horizontal and vertical flips, and correspondingly update other related memory information.

To automatically generate the task instruction from the human demonstration video, we prompt GPT-4V \cite{openai2023gpt4v} to describe the video in natural language, as shown in Fig. \ref{fig:memory}.

\subsection{Retrieval}

When humans encounter a new object to manipulate, they retrieve the most similar experience, in terms of semantics and geometry, from their memory \cite{jeannerod1995graspingretri}. Inspired by this, we propose a two-stage method that first performs semantic retrieval followed by geometric retrieval.

\textbf{Semantic Retrieval.} To consider the semantic information of objects and tasks during retrieval, we draw inspiration from the Retrieval-Augmented Generation (RAG) system \cite{lewis2020RAG} and construct a semantic retrieval module. This module consists of two steps: (1) performing coarse semantic retrieval, which is fast but only considers coarse semantic similarity, and (2) refining the retrieval with fine-grained semantic similarity using the semantic reasoning capabilities of VLMs. Specifically, we first use CLIP \cite{radford2021CLIP} to encode the semantic information of object images and task instructions from the memory into feature vectors. During inference, the module coarsely retrieves relevant memory instances based on the input image and the task. Then, leveraging the prior knowledge of foundation models, we employ GPT-4V to refine this selection by evaluating the candidates against task instructions and object semantics and select the Top-k candidates.

\textbf{Geometric Retrieval.} Recent research \cite{tang2023emergentDIFT} \cite{zhang2024dino+dift} shows that deep dense feature maps generated by vision foundation models contain rich information for precise dense matching. However, as shown in \cite{zhang2024geo}, these models are vulnerable to viewpoint changes. Therefore, to achieve robust dense matching in the later stage, we perform a geometric retrieval to select a candidate most similar to the current viewpoint of the target object. Specifically, we calculate the Instance Matching Distance (IMD) \cite{zhang2024geo} between the semantically retrieved candidates and the target object, based on the fused features of Stable Diffusion and DINOv2 (SD+DINOv2) as in \cite{zhang2024dino+dift}, to identify the candidate memory instance with the most similar viewpoint.

\subsection{Transfer}

After retrieving the optimal candidate from the memory, we transfer the grasping experience, including the TOG position and direction constraints, to the target object image $I_B$ through semantic correspondence. The transfer of $\mathbf{p}_{A}$ is relatively straightforward, as we can directly map $\mathbf{p}_{A}$ to $I_B$ using the dense feature mapping capability of vision foundation models SD+DINOv2 \cite{zhang2024dino+dift}. For the transfer of $\mathbf{v}_{A}$, we need to establish the relative pose relationship between the demonstration object and the target object. Establishing the pose relationship involves two steps: (1) performing 2D-2D matching between the two objects and (2) constructing 2D-3D correspondences based on the 2D-2D matching, followed by solving for the relative pose using the PnP algorithm \cite{lepetit2009epnp}. Specifically, using the object masks obtained by SAM \cite{kirillov2023sam}, we utilize the deep features of SD+DINOv2 and apply the Best Buddies Nearest Neighbour matching \cite{amir2021bbs} to construct 2D-2D correspondences $\mathcal{M}_{2D}$ between the two objects. Notably, due to the powerful semantic correspondence capabilities of SD+DINOv2's deep features, the transfer module achieves strong cross-category generalization. Based on $\mathcal{M}_{2D}$ and the target object's point cloud $P_c$, we construct 2D-3D correspondences $\mathcal{M}_{3D}=\{C_1,C_2\}$, where $C_1 = \{[u_{i}, v_{i}]\}_{i=1}^N$ and $C_2 = \{[x_{i}, y_{i}, z_{i}]\}_{i=1}^N$ (the 3D points are obtained by extracting the depth information for each pixel from the RGB-D camera). Finally, by solving the PnP problem in $\mathcal{M}_{3D}$, we calculate the relative pose $\mathbf{T_A} = [\mathbf{R}_A~\mathbf{t}_A]$ between the demonstration object in $I_A$ and the target object in $I_B$, as shown in:
\begin{equation}
    \mathbf{R}_A, \mathbf{t}_A = \underset{\mathbf{R}, \mathbf{t}}{\arg\min} \sum_{i=1}^{N} \left\| \begin{bmatrix} u_{i} \\ v_{i}  \end{bmatrix} - \boldsymbol{\pi} \left( \mathbf{R} \begin{bmatrix} x_{i} \\ y_{i} \\ z_{i} \end{bmatrix} + \mathbf{t} \right) \right\|_2 ,
\end{equation}
where the projection function $\boldsymbol{\pi}(\cdot)$ represents the mapping of the demonstration camera's intrinsic matrix $\mathbf{K}_A$ with normalized coordinates. With $\mathbf{T}_{A}$, we can transfer $\mathbf{v}_{A}$ from the memory to $I_B$ to obtain $\mathbf{v}_{B}$.

\subsection{Alignment}

In the final stage, RTAGrasp aligns the transferred constraints, $\mathbf{p}_{B}$ and $\mathbf{v}_{B}$, to a task-oriented grasp pose that a robot can directly execute. Here, we adopt a sampling-and-evaluation approach. Specifically, we first use a task-agnostic grasp sampler to generate a set of stable grasp candidates on the target object. The sampler takes the point cloud of the target object as input and outputs a set of stable 6-DOF grasp poses $\left\{\mathbf{R}=[\mathbf{o}_x~\mathbf{o}_y~\mathbf{o}_z]  \in \mathbb{R}^{3 \times 3}, \mathbf{t} \in \mathbb{R}^{3 \times 1} \right\}$. Each grasp pose is with a stability score $\mathcal{S}_{\text{geo}}$. We then calculate a task-compatibility score $\mathcal{S}_{\text{task}}$ for each grasp candidate with the TOG constraints, as illustrated in:
\begin{equation}
     \mathcal{S}_{\text{task}}=\frac{\mathbf{v}_{B} \cdot \mathbf{o}_z}{\|\mathbf{v}_{B}\| \|\mathbf{o}_z\|} + \exp\left(-\frac{\|\mathbf{t} - \mathbf{p}_{B}\|^2}{2\sigma^2}\right) \label{eq:task},
\end{equation}
where $\sigma=0.1$. The cosine similarity measures the alignment between the grasp candidate's direction and the TOG constraint, while the Gaussian decay function captures the positional deviation. We compute the final score for each grasp candidate as $\mathcal{S}=0.95\mathcal{S}_{\text{task}}+0.05\mathcal{S}_{\text{geo}}$ prioritizing task-compatibility over stability since most candidates are stable. The sampling-and-evaluation approach allows us to leverage the outcomes from prior work in stable grasp generation while avoiding the intricate retargeting from the human hand to the robot gripper. In our implementation, we use Contact-GraspNet \cite{sundermeyer2021contactgraspnet} as the grasp sampler, but other stable samplers would also work. The robot then selects the candidate with the highest score for execution.

\section{Experiments}

In this section, we compare RTAGrasp with existing training-based and training-free methods and conduct extensive ablation studies. Furthermore, we evaluate the practicality of RTAGrasp through real-world experiments.

\begin{table*}[t]
\caption{Quantitative Results of Task-oriented Grasp Experiments on the Dataset}
\label{tab:grasp_quantitative}
\renewcommand\arraystretch{1.8}
\setlength\tabcolsep{3pt} 
\centering
\vspace*{-0.1in}
\begin{tabular}{lccccccccccccccc}
\toprule
\multicolumn{1}{c}{\multirow{2}{*}{\textbf{Setting}}} & \multicolumn{1}{c}{\multirow{2}{*}{\textbf{Method}}} & \multicolumn{12}{c}{\textbf{Unseen Categories Success Rate}} &\multicolumn{1}{c}{\multirow{2}{*}{\textbf{Total}}} \\ 
\cline{3-14}
& & \textbf{Scissors} & \textbf{Bottle} & \textbf{Trowel} & \textbf{Ladle} & \textbf{Masher} & \textbf{Brush} & \textbf{Opener} & \textbf{Peeler}  & \textbf{Mortar}& \textbf{Clamp}& \textbf{Roller}& \textbf{Skimmer}\\ 
\toprule
\textbf{Training-based} & GraspGPT & 63.00\% & 62.00\% & 53.00\% & 83.00\% & 71.00\% & 72.00\% & 62.00\% & \textbf{75.00}\% & 62.00\% & 76.00\% & 67.00\% & 65.00\% & 67.58\%  \\ 
\hline 
\multirow{3}{*}{\textbf{Training-free}} & Lan-grasp & 60.00\% & 61.00\% & 83.00\% & 85.00\% & 63.00\% & 74.00\% & 70.00\% & 72.00\% & 79.00\%& 84.00\% & 75.00\% & 73.00\% & 73.25\% \\
                                        & RoboABC  & 73.00\% & 43.00\% & 50.00\% & 81.00\% & 57.00\% & 68.00\% & 67.00\% & 45.00\% & 67.00\% & 86.00\% & 73.00\% & 80.00\% & 65.83\% \\
                                        & Ours     & \textbf{79.00}\% & \textbf{77.00}\% & \textbf{86.00}\% & \textbf{88.00}\% & \textbf{72.00}\% & \textbf{81.00}\% & \textbf{78.00}\% & 74.00\% & \textbf{83.00}\% & \textbf{89.00}\% & \textbf{76.00}\% & \textbf{83.00}\% & \textbf{80.50}\%\\ 
\bottomrule
\end{tabular}
\vspace*{-0.2in}
\end{table*}

\subsection{Perception Experiments}

We conduct perception experiments on the public TOG benchmark, TaskGrasp dataset. In these experiments, ``seen categories" refer to object categories used for training (training-based) or stored in memory (training-free), while ``unseen categories" refer to those not used. For all baselines, we test on 12 unseen and 48 seen categories, comprising a total of 190 instances, with 100 trials conducted per instance. The Top-1 success rate is used as the evaluation metric.
Detailed category names are provided in the supplementary material. Since the TaskGrasp dataset only includes manual TOG annotations and lacks the human TOG demonstrations, we collect 64 demonstrations, covering 48 seen categories, to construct a memory. 

We compare our method with three baselines:

\textbf{GraspGPT} \cite{tang2023graspgpt} is the SOTA training-based method that transfers task-oriented grasps from seen to unseen categories using semantic relationships learned from the TaskGrasp dataset. Following the original data split, we train GraspGPT on 48 seen categories and test on 12 unseen categories.

\textbf{Lan-grasp} \cite{mirjalili2023langrasp} is a training-free method that leverages foundation models for 2D TOG region localization. Since it only identifies coarse regions, we randomly select grasp poses within those regions.

\textbf{RoboABC} \cite{ju2024roboABC} is a training-free method based on retrieval and transfer for grasping point prediction. It does not consider task constraints (i.e., different tasks might correspond to different grasping points) when transferring grasping points and does not address the problem of ``how to grasp".

We first compare RTAGrasp with the training-based method. As shown by the experimental results in Table \ref{tab:grasp_quantitative}, our method achieves an average success rate of 80.50\% on unseen categories, outperforming GraspGPT by 12.92\%. This results demonstrate that using the dense semantic matching capabilities of foundation models for TOG constraints transfer offers better generalization compared to those who transfer grasps based on knowledge learned on a manually annotated dataset. Then, we conduct experiments on seen categories. As shown in Table \ref{tab:exe3}, RTAGrasp achieves a success rate of 91.17\% using only 64 demonstration videos from the seen categories. Although GraspGPT achieves a higher success rate of 97.40\% on the seen categories, it relies on a dataset with over 30,000 manual TOG annotations on 190 instances for training (compared to only 64 demonstrations). With three orders of magnitude less data, RTAGrasp performs only slightly worse than GraspGPT on seen categories, highlighting the data efficiency of our approach. In conclusion, the comparison with the SOTA training-based method demonstrates that RTAGrasp achieves competitive performance with minimal manual efforts, while offering better generalization and higher data efficiency.

\vspace*{-0.1in}
\begin{table}[h]
\caption{Quantitative Comparison with Training-based Method}
\label{tab:exe3}
\renewcommand\arraystretch{1.8}
\setlength\tabcolsep{2.5pt} 
\centering
\vspace*{-0.1in}
\begin{tabular}{lccccc}
\toprule
\textbf{Method} & \textbf{Seen Categories} & \textbf{Unseen Categories} & \textbf{Data Volume}\\
\hline 
GraspGPT &\textbf{97.40\% }& 67.58\% & Over 30,000 samples\\
Ours & 91.17\% & \textbf{80.50\%} & 64 samples\\
\bottomrule
\end{tabular}
\end{table}

We then compare RTAGrasp with training-free methods on both seen and unseen categories. For Lan-grasp \cite{mirjalili2023langrasp}, since it directly leverages the common sense knowledge from foundation models to predict TOG regions, not requiring any demonstration, all categories are considered as unseen categories. As shown in Table \ref{tab:exe2}, our method consistently outperforms existing training-free methods. The comparison with existing training-free methods demonstrates that (1) foundation models cannot directly provide full knowledge for TOG, and (2) extracting complete TOG constraints from human demonstrations is crucial for TOG.

\vspace*{-0.1in}
\begin{table}[h]
\caption{Quantitative Comparison with Training-free Methods}
\label{tab:exe2}
\renewcommand\arraystretch{1.8}
\setlength\tabcolsep{8pt} 
\centering
\vspace*{-0.1in}
\begin{tabular}{lccccc}
\toprule
\textbf{Method} & \textbf{Seen Categories} & \textbf{Unseen Categories} \\
\hline 
Lan-grasp &72.42\% & 73.25\% \\
RoboABC & 70.08\% & 65.83\% \\
Ours & \textbf{91.17\%} & \textbf{80.50\%} \\
\bottomrule
\end{tabular}
\vspace*{-0.2in}
\end{table}

\subsection{Ablation Study}

To further evaluate the rationale behind our framework design, we conduct extensive ablation studies on each component, using success rate as the evaluation metric. Experiments are performed on both seen and unseen categories.

\textbf{Ablation on Retrieval Module:} To validate the contribution of the semantic retrieval module, we replace it with a method that calculates cosine similarity after concatenating encoded task text features and image features. As shown in the Table \ref{tab:ablation},  success rates drop by 6.75\% and 3.75\%, highlighting the necessity of our semantic retrieval module. Moreover, removing the geometric retrieval module results in a success rate drop of over 10\%. A significant difference in the object's viewpoint can greatly affect the accuracy of subsequent matching. These results emphasize the effectiveness of both retrieval modules in RTAGrasp.

\begin{figure*}[htbp]
  \centering
  \begin{tikzpicture}[inner sep = 0pt, outer sep = 0pt]
    \node[anchor=south west] (fnC) at (0in,0in)
      {\includegraphics[height=3.5in,clip=true,trim=0in 0in 0in 0in]{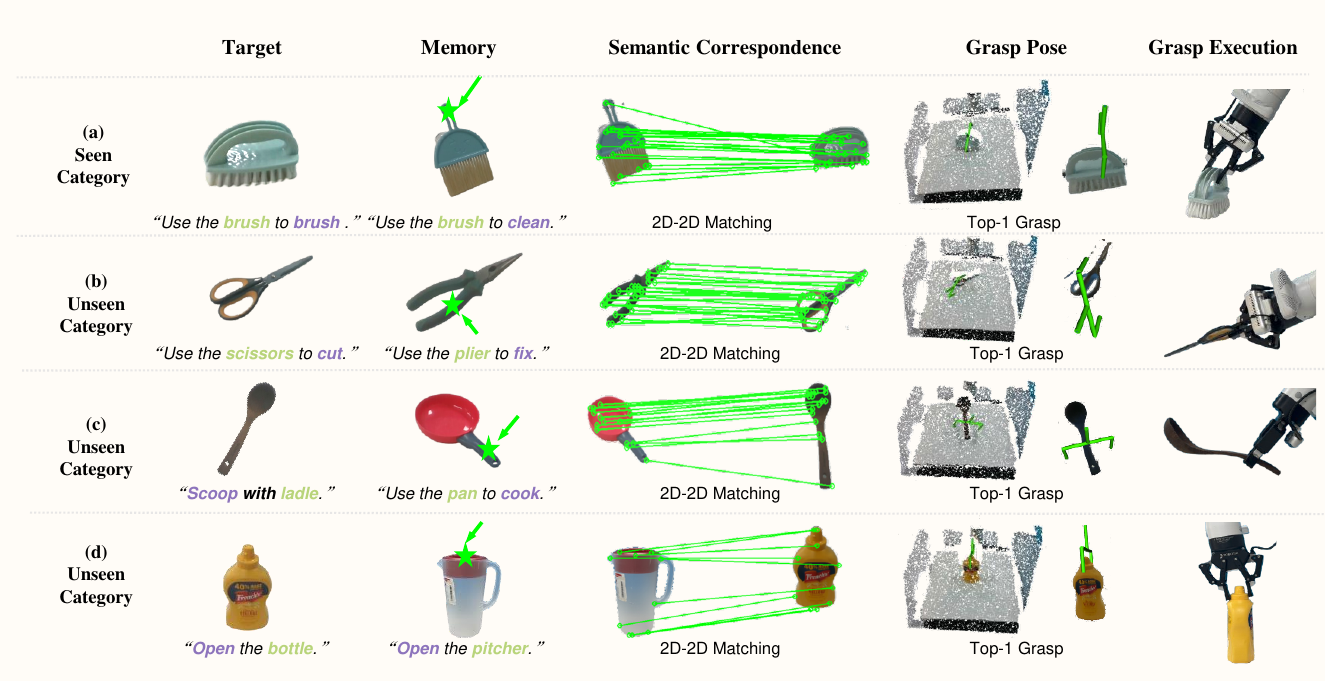}};
  \end{tikzpicture}
  \caption{Qualitative results of TOG. Each row is a visualization of the intermediate results for an object in an experimental scene.}
  \label{fig:exe}
  \vspace*{-0.2in}
\end{figure*}

\vspace*{-0.1in}
\begin{table}[h]
\caption{Ablation Study Results}
\label{tab:ablation}
\renewcommand\arraystretch{1.8}
\setlength\tabcolsep{7pt} 
\centering
\vspace*{-0.1in}
\begin{tabular}{lccccc}
\toprule
\textbf{Ablated Version} & \textbf{Seen Categories} & \textbf{Unseen Categories} \\
\hline 
RAG $\rightarrow$ Cos. Sim. & 84.42\%& 76.75\% \\
w/o Geom.Rtrvl &80.83\% & 66.92\% \\
\hline 
SD+DINOv2 $\rightarrow$ CLIP & 70.25\% & 42.33\% \\
SD+DINOv2 $\rightarrow$ DINOv2 & 78.50\% & 58.58\% \\
SD+DINOv2 $\rightarrow$ SD & 86.33\% & 74.58\% \\
\hline 
\textbf{Full Pipeline} &\textbf{91.17\%} & \textbf{80.50\%} \\
\bottomrule
\end{tabular}
\vspace*{-0.1in}
\end{table}

\textbf{Ablation on Transfer Module:} We experiment with replacing the SD+DINOv2 \cite{zhang2024dino+dift} model in the transfer module with other vision foundation models. As shown in Table \ref{tab:ablation}, SD+DINOv2 significantly outperforms CLIP \cite{radford2021CLIP} and DINOv2 \cite{oquab2023dinov2} on both seen and unseen categories and slightly surpasses SD \cite{rombach2022SIFT} alone. Therefore, we select SD+DINOv2 as the transfer module for the final pipeline.

\begin{figure}[th]
  \centering
  \vspace*{-0.1in}
  \begin{tikzpicture}[inner sep = 0pt, outer sep = 0pt]
    \node[anchor=south west] (fnC) at (0in,0in)
      {\includegraphics[height=1.8in,clip=true,trim=0in 0in 0in 0in]{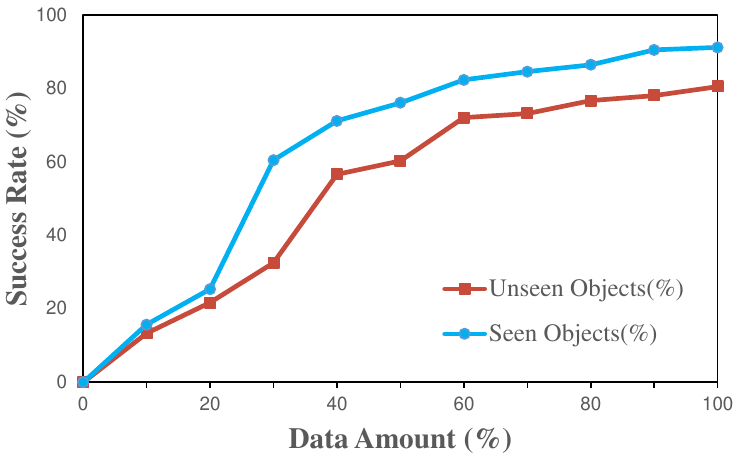}};
  \end{tikzpicture}
    \vspace*{-0.1in}
  \caption{Performances with different data amounts for retrieval.}
  \label{fig:data_amount}
  \vspace*{-0.2in}
\end{figure}

\textbf{Effect of Data Volume:} We examine the effect of demonstration data volume on the performance of RTAGrasp, as illustrated in Fig. \ref{fig:data_amount}. Overall, the results suggest that RTAGrasp's performance improves with the increase in data volume. This aligns with our intuition. Interestingly, when the data volume is reduced to 60\%, performance remains comparable to that achieved with the full dataset, highlighting the data efficiency of our approach.

\FloatBarrier

\subsection{Real-World Experiments}

In real-world experiments, we deploy RTAGrasp on a Kinova Gen3 robot arm equipped with a RealSense D435i camera mounted on its wrist. The setup follows the perception experiment, testing on 12 unseen categories with 24 instances in total. Each object instance is tested five times with random placements. Consistent with previous work \cite{tang2023graspgpt}, the evaluation of real-world experiments is divided into three stages: perception, planning, and action. As shown in Table \ref{tab:real_exe}, RTAGrasp achieves a 73.33\% overall success rate, outperforming baseline methods across all phases. This confirms its effectiveness in practical applications. Some qualitative results are provided in Fig. \ref{fig:exe}. 

\vspace{-0.1in}
\begin{table}[h]
\caption{Quantitative Results of Real-world Experiments}
\label{tab:real_exe}
\centering
\renewcommand\arraystretch{1.5}
\setlength\tabcolsep{8pt} 
\vspace*{-0.1in}
\begin{tabular}{lcccc}
\toprule
\multicolumn{1}{c}{\multirow{2}{*}{\textbf{Method}}} & \multicolumn{3}{c}{\textbf{Unseen Categories Performance}} & \multicolumn{1}{c}{\multirow{2}{*}{\textbf{Success}}} \\ \cline{2-4} 
& \textbf{Perception} & \textbf{Planning} & \textbf{Action}  \\ \toprule
GraspGPT & 81/120 &  79/120 &  73/120 & 60.83\% \\
Lan-grasp & 86/120 &  80/120 &  78/120 & 65.00\% \\
RoboABC & 70/120 &  67/120 &  65/120 & 54.17\% \\
Ours & 94/120 &  90/120 &  88/120 & \textbf{73.33\%} \\ \bottomrule
\end{tabular}
\vspace*{-0.1in}
\end{table}

We use the grasps generated by RTAGrasp as the first step of the manipulation tasks and integrated existing frameworks to complete tests on several tasks. The experiment videos are available on our \href{https://sites.google.com/view/rtagrasp/home}{project website}.

\section{Conclusions}

In this work, we effortlessly extract complete TOG constraints from human demonstration videos and then explicitly transfer them to robots with RTAGrasp, a Retrieval, Transfer, and Alignment framework that leverages the semantic correspondence capabilities of vision foundation models. Experimental results demonstrate that our method outperforms existing TOG approaches on both seen and unseen object categories, and can be effectively deployed in real-world robotic applications.

In the future, we aim to build a large-scale robotic memory as a TOG foundation model and incorporate it into downstream manipulation systems. Additionally, we plan to perform auto-augmentation over the collected memory, enabling life-long self-improvement through continual learning and the integration of new grasping experiences.



\bibliographystyle{IEEEtran}
\balance
\bibliography{root}

\end{document}